\documentclass{article} 
\usepackage{nips13submit_e,times}
\usepackage{hyperref}
\usepackage{url}
\usepackage{graphicx}
\usepackage{amsmath}
\usepackage{amsfonts}
\usepackage{mathrsfs}
\usepackage{caption}

\usepackage{wrapfig}
\usepackage{algpseudocode}
\usepackage{algorithm}
\usepackage{algorithmicx}

\newcommand{\cX}{{\bf X}}
\newcommand{\cW}{{\bf \mathsf{W}}}
\newcommand{\cZ}{{\bf Z}}
\newcommand{\cQ}{{\bf Q}}
\newcommand{\sw}{{\bf w}}
\newcommand{\ow}{{\bf w}^{(out)}}
\newcommand{\cL}{\mathcal{L}}
\newcommand{\sL}{\ell}
\newcommand{\cg}{{\bf g}}
\newcommand{\sg}{{\bf \hat{g}}}
\newcommand{\cE}{{\mathbb{E}}}
\newcommand{\aP}{{\mathcal{P}}}
\newcommand{\aQ}{{\mathcal{Q}}}
\newcommand{\gp}{{\bf gp}}
\newcommand{\gq}{{\bf gq}}
\newcommand{\gf}{{\bf gf}}
\newcommand{\cWs}{{\bf \mathsf{W}}^{\star}}

\nipsfinalcopy 
\newtheorem{theor}{Theorem}
\newtheorem{lemma}{Lemma}
\newenvironment{proof}[1][Proof]{\begin{trivlist}
\item[\hskip \labelsep {\bfseries #1}]}{\end{trivlist}}
\newenvironment{definition}[1][Definition]{\begin{trivlist}
\item[\hskip \labelsep {\bfseries #1}]}{\end{trivlist}}

\newenvironment{remark}[1][Remark]{\begin{trivlist}
\item[\hskip \labelsep {\bfseries #1}]}{\end{trivlist}}

\newcommand{\qed}{\nobreak \ifvmode \relax \else
      \ifdim\lastskip<1.5em \hskip-\lastskip
      \hskip1.5em plus0em minus0.5em \fi \nobreak
      \vrule height0.75em width0.5em depth0.25em\fi}

\title{Deeply-Supervised Nets}

\author{
Chen-Yu Lee \thanks{equal contribution} \\
Dept. of EECS, UCSD\\
\texttt{chl260@ucsd.edu} \\
\And
Saining Xie $^{*}$ \\
Dept. of CSE and CogSci, UCSD\\
\texttt{s9xie@ucsd.edu} \\
\AND
Patrick Gallagher \\
Dept. of CogSci, UCSD\\
\texttt{rexaran@gmail.com} \\
\And
Zhengyou Zhang \\
Microsoft Research \\
\texttt{zhang@microsoft.com } \\
\And
Zhuowen Tu \thanks{Corresponding author. Patent disclosure, UCSD Docket No. SD2014-313, filed on May 22, 2014.} \\
Dept. of CogSci, UCSD\\
\texttt{ztu@ucsd.edu} \\
}
\begin{document}

\maketitle

\begin{abstract}
\vspace{-5mm}
Our proposed deeply-supervised nets (DSN) method simultaneously minimizes classification error while making the learning process of hidden layers direct and transparent. We make an attempt to boost the classification performance by studying a new formulation in deep networks. Three aspects in convolutional neural networks (CNN) style architectures are being looked at: (1) transparency of the intermediate layers to the overall classification; (2) discriminativeness and robustness of learned features, especially in the early layers; (3) effectiveness in training due to the presence of the exploding and vanishing gradients.  We introduce ``companion objective'' to the individual hidden layers, in addition to the overall objective at the output layer (a different strategy to layer-wise pre-training). We extend techniques from stochastic gradient methods to analyze our algorithm. The advantage of our method is evident and our experimental result on benchmark datasets shows significant performance gain over existing methods (e.g. all state-of-the-art results on MNIST, CIFAR-10, CIFAR-100, and SVHN).
\end{abstract}

\section{Introduction}
Much attention has been given to a resurgence of neural networks, deep learning (DL) in particular, which can be of unsupervised \cite{Hinton06}, supervised \cite{CNN}, or a hybrid form \cite{Lee09}. Significant performance gain has been observed, especially in the presence of large amount of training data, when deep learning techniques are used for image classification \cite{dropout,Le10} and speech recognition \cite{Dahl12}. On the one hand, hierarchical and recursive networks \cite{Elman91,Hinton06,CNN} have demonstrated great promise in automatically learning thousands or even millions of features for pattern recognition; on the other hand concerns about deep learning have been raised and many fundamental questions remain open.

Some potential problems with the current DL frameworks include: reduced transparency and discriminativeness of the features learned at hidden layers \cite{Zeiler13}; training difficulty due to exploding and vanishing gradients \cite{Glorot10,Pascanu14}; lack of a thorough mathematical understanding about the algorithmic behavior, despite of some attempts made on the theoretical side \cite{Eigen14}; dependence on the availability of large amount of training data \cite{dropout}; complexity of manual tuning during training \cite{imagenet}. Nevertheless, DL is capable of automatically learning and fusing rich hierarchical features in an integrated framework.
Recent activities in open-sourcing and experience sharing \cite{dropout,DeCAF,theano} have also greatly helped the adopting and advancing of DL in the machine learning community and beyond. Several techniques, such as dropout \cite{dropout}, dropconnect \cite{dropcon}, pre-training \cite{Dahl12}, and data augmentation \cite{mulCOL}, have been proposed to enhance the performance of DL from various angles, in addition to a variety of engineering tricks used to fine-tune feature scale, step size, and convergence rate. Features learned automatically by the CNN algorithm \cite{CNN} are intuitive \cite{Zeiler13}. Some portion of features, especially for those in the early layers, also demonstrate certain degree of opacity \cite{Zeiler13}. This finding is also consistent with an observation that different initializations of the feature learning at the early layers make negligible difference  to the final classification \cite{Dahl12}. In addition, the presence of vanishing gradients also makes the DL training slow and ineffective \cite{Glorot10}. In this paper, we address the feature learning problem in DL by presenting a new algorithm, deeply-supervised nets (DSN), which enforces direct and early supervision for both the hidden layers and the output layer. We introduce {\em companion objective} to the individual hidden layers, which is used as an additional constraint (or a new regularization) to the learning process. 
Our new formulation significantly enhances the performance of existing supervised DL methods. We also make an attempt to provide justification for our formulation using stochastic gradient techniques. We show an improvement of the convergence rate of the proposed method over standard ones, assuming local strong convexity of the optimization function (a very loose assumption but pointing to a promising direction).

Several existing approaches are particularly worth mentioning and comparing with. In \cite{pretrain}, layer-wise supervised pre-training is performed. Our proposed method does not perform pre-training and it emphasizes the importance of minimizing the output classification error while reducing the prediction error of each individual layer. This is important as the backpropagation is performed altogether in an integrated framework.
In \cite{Snoek12}, label information is used for unsupervised learning. Semi-supervised learning is carried in deep learning \cite{Weston12}.
In \cite{Tang13}, an SVM classifier is used for the output layer, instead of the standard softmax function in the CNN \cite{CNN}. Our framework (DSN), with the choice of using SVM, softmax or other classifiers, emphasizes the direct supervision of each intermediate layer. In the experiments, we show consistent improvement of DSN-SVM and DSN-Softmax over CNN-SVM and CNN-Softmax respectively. 
We observe all state-of-the-art results on MNIST, CIFAR-10, CIFAR-100, and SVHN. It is also worth mentioning that our formulation is inclusive to various techniques proposed recently such as averaging \cite{mulCOL}, dropconnect \cite{dropcon}, and Maxout \cite{maxout}. We expect to see more classification error reduction with careful engineering for DSN.

\section{Deeply-Supervised Nets}
In this section, we give the main formulation of the proposed deeply-supervised nets (DSN). We focus on building our infrastructure around supervised CNN style frameworks \cite{CNN,DeCAF,theano} by introducing classifier, e.g. SVM model \cite{vapnik}, to each layer. An early attempt to combine SVM with DL was made in \cite{Tang13}, which however has a different motivation with ours and only studies the output layer with some preliminary experimental results.

\subsection{Motivation}

\begin{figure*}[t]
\begin{center}
\includegraphics[width=0.8\linewidth]{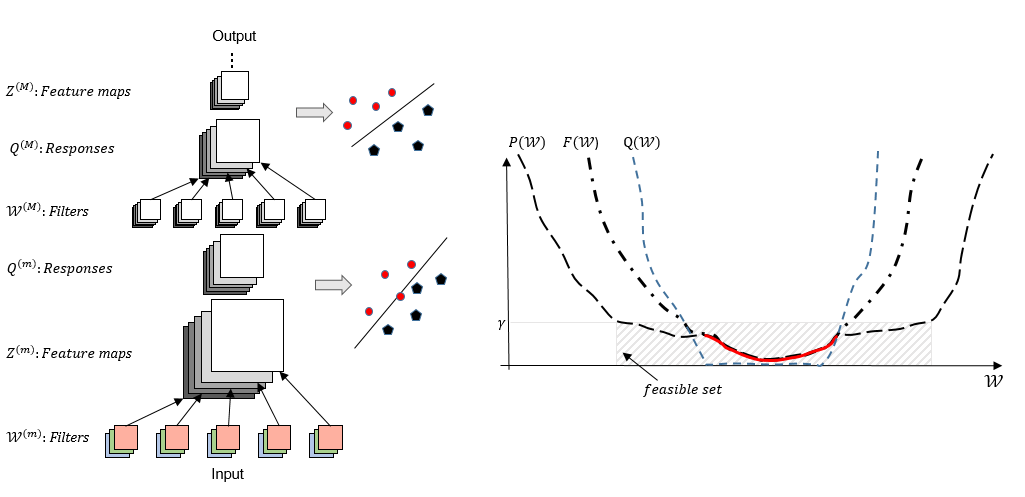}
\begin{tabular}{ll}
(a) DSN illustration  & \hspace{4cm} (b) functions\\
\end{tabular}
\end{center}
\caption{Network architecture for the proposed deeply-supervised nets (DSN).}
\label{fig:architecture}
\vspace{-5mm}
\end{figure*}

We are motivated by the following simple observation: in general, a discriminative classifier trained on highly discriminative features will display better performance than a discriminative classifier trained on less discriminative features. If the features in question are the hidden layer feature maps of a deep network, this observation means that the performance of a discriminative classifier trained using these hidden layer feature maps can serve as a proxy for the quality/discriminativeness of those hidden layer feature maps, and further to the quality of the upper layer feature maps.
By making appropriate use of this feature quality feedback at each hidden layer of the network, we are able to directly influence the hidden layer weight/filter update process to favor highly discriminative feature maps. This is a source of supervision that acts deep within the network at each layer; when our proxy for feature quality is good, we expect to much more rapidly approach the region of good features than would be the case if we had to rely on the gradual backpropagation from the output layer alone.
We also expect to alleviate the common problem of having gradients that ``explode'' or ``vanish''.
One concern with a direct pursuit of feature discriminativeness at all hidden layers is that this might interfere with the overall network performance, since it is ultimately the feature maps at the output layer which are used for the final classification; our experimental results indicate that this is not the case.


Our basic network architecture will be similar to the standard one used in the CNN framework. Our additional deep feedback is brought in by associating a companion local output with each hidden layer.  We may think of this companion local output as analogous to the final output that a truncated network would have produced.
Backpropagation of error now proceeds as usual, with the crucial difference that we now backpropagate not only from the final layer but also simultaneously from our local companion output. The empirical result suggests the following main properties of the companion objective: {(1) it acts as a kind of feature regularization (although an unusual one), which leads to significant reduction to the testing error but not necessarily to the train error; (2) it results in faster convergence, especially in presence of small training data (see Figure (\ref{fig:mnist}) for an illustration on a running example).}

\subsection{Formulation}

We focus on the supervised learning case and let $S=\{(\cX_i, y_i),i=1..N\}$ be our set of input training data where sample $\cX_i \in R^n$ denotes the raw input data and $y_i \in \{1,..,K\}$ is the corresponding groundtruth label for sample $X_i$. We drop $i$ for notational simplicity, since each sample is considered independently. The goal of deep nets, specifically convolutional neural networks (CNN) \cite{CNN}, is to learn layers of filters and weights for the minimization of classification error at the output layer. Here, we absorb the bias term into the weight parameters and do not differentiate weights from filters and denote a recursive function for each layer $m=1..M$ as:
\begin{eqnarray}
	& & \cZ^{(m)} = f(\cQ^{(m)}), \;\; 	and \quad \cZ^{(0)}\equiv \cX,\label{eq:Z} \\
  & & \cQ^{(m)} = \cW^{(m)} * \cZ^{(m-1)}. \label{eq:Q}
\end{eqnarray}
$M$ denotes the total number of layers; $\cW^{(m)}, m=1..M$ are the filters/weights to be learned; $\cZ^{(m-1)}$ is the feature map produced at layer $m-1$; $\cQ^{(m)}$ refers to the convolved/filtered responses on the previous feature map; $f()$ is a pooling function on $\cQ$; 
Combining all layers of weights gives
\[
   \cW = (\cW^{(1)},...,\cW^{(M)}).
\]
Now we introduce a set of classifiers, e.g. SVM (other classifiers like Softmax can be applied and we will show results using both SVM and Softmax in the experiments), one for each hidden layer,
\[
   \sw = (\sw^{(1)},...,\sw^{(M-1)}),
\]
in addition to the $\cW$ in the standard CNN framework.
We denote the $\ow$ as the SVM weights for the output layer. 
Thus, we build our overall combined objective function as:
\begin{equation}
{\lVert \ow \rVert^2 + \cL(\cW, \ow) + \sum_{m=1}^{M-1} \alpha_m  [\lVert \sw^{(m)} \rVert^2 + \sL(\cW, \sw^{(m)}) - \gamma]_+,}
\label{eq:total2}
\end{equation}
where
\begin{equation}
\label{eq:layer}
{\cL(\cW, \sw^{(out)})  = \sum_{y_k \ne y} [1- <\sw^{(out)},  \phi(\cZ^{(M)}, y) - \phi(\cZ^{(M)}, y_k)>]_{+}^2 }
\end{equation}
and
\begin{equation}
\label{eq:closs}
{\sL(\cW, \sw^{(m)})  = \sum_{y_k \ne y} [1- <\sw^{(m)}, \phi(\cZ^{(m)}, y) - \phi(\cZ^{(m)}, y_k)>]_{+}^2 }
\end{equation}

We name $\cL(\cW, \sw^{(M)})$ as the {\em overall loss} (output layer) and $\sL(\cW, \sw^{(m)})$  as the {\em companion loss} (hidden layers), which are both squared hinge losses of the prediction errors. 
The above formulation can be understood intuitively: in addition to learning convolution kernels and weights, $\cW^{\star}$, as in the standard CNN model \cite{CNN}, enforcing a constraint at each hidden layer for directly making a good label prediction gives a strong push for having discriminative and sensible features at each individual layer.
In eqn. (\ref{eq:total2}), $\lVert \sw^{(out)} \rVert^2$ and $\cL(\cW, \sw^{(out)})$
are respectively the margin and squared hinge loss of the SVM classifier (L2SVM \footnote{It makes negligible difference between L1SVM and L2SVM.}) at the output layer (we omit the balance term $C$ in front of the hinge for notational simplicity);
in eqn. (\ref{eq:layer}), $\lVert \sw^{(m)} \rVert^2$ and $\sL(\cW, \sw^{(m)})$ are respectively the margin and squared hinge loss of the SVM classifier at each hidden layer. Note that for each $\sL(\cW, \sw^{(m)})$,
the $\sw^{(m)}$ directly depends on $\cZ^{(m)}$, which is dependent on $\cW^{1},..,\cW^{m}$ up to the $m$th layer. $\cL(\cW, \sw^{(out)})$ depends on $\sw^{(out)}$, which is decided by the entire $\cW$. The second term  in eqn. (\ref{eq:total2}) often goes to zero during the course of training; this way, the overall goal of producing good classification of the output layer is not altered and the companion objective just acts as a proxy or regularization.  This is achieved by having $\gamma$ as a threshold (a hyper parameter) in the second term of eqn. (\ref{eq:total2}) with a hinge loss: once the overall value of the hidden layer reaches or is below $\gamma$, it vanishes and no longer plays role in the learning process.  $\alpha_m$ balances the importance of the error in the output objective and the companion objective. In addition, we could use a simple decay function as $\alpha_m \times 0.1 \times (1-t/N) \rightarrow \alpha_m$ to enforce the second term to vanish after certain number of iterations, where $t$ is the epoch step and $N$ is the total number of epochs (wheather or not to have the decay on $\alpha_m$ might vary in different experiments although the differences may not be very big).

To summarize, we describe this optimization problem as follows: we want to learn filters/weights $\cW$ for the entire network such that an SVM classifier $\sw^{(out)}$ trained on the output feature maps (that depend on those filters/features) will display good performance. We seek this output performance while also requiring some ``satisfactory'' level of performance on the part of the hidden layer classifiers. We are saying: restrict attention to the parts of feature space that, when considered at the internal layers, lead to highly discriminative hidden layer feature maps (as measured via our proxy of hidden-layer classifier performance). The main difference between eqn. (\ref{eq:total2}) and previous attempts in layer-wise supervised training is that we perform the optimization altogether with a robust measure (or regularization) of the hidden layer. For example, greedy layer-wise pretraining was performed as either initialization or fine-tuning which results in some overfitting \cite{pretrain}. The state-of-the-art benchmark results demonstrate the particular advantage of our formulation. As shown in Figure \ref{fig:mnist}(c), indeed both CNN and DSN reach training error near zero but DSN demonstrates a clear advantage of having a better generalization capability. 


To train the DSN model using SGD, the gradients of the objective function w.r.t the parameters in the model are:
\begin{equation} \label{eq:grad}
\begin{split}
& {\scriptstyle \frac{\partial F}{\partial \sw^{(out)}} = 2\sw^{(out)} - 2 \sum_{y_k \ne y} [\phi(\cZ^{(M)}, y) - \phi(\cZ^{(M)}, y_k)][1- <\sw^{(out)},  \phi(\cZ^{(M)}, y) - \phi(\cZ^{(M)}, y_k)>]_{+}} \\
& {\scriptstyle \frac{\partial F}{\partial \sw^{(m)}}} = \begin{cases} {\scriptstyle \alpha_m \left\{ 2\sw^{(m)} - 2 \sum_{y_k \ne y} [\phi(\cZ^{(m)}, y) - \phi(\cZ^{(m)}, y_k)][1- <\sw^{(m)},  \phi(\cZ^{(m)}, y) - \phi(\cZ^{(m)}, y_k)>]_{+} \right\} },
\quad \text{otherwise} \\
0, \qquad\qquad\qquad\qquad\qquad\qquad\qquad\qquad\qquad\qquad\qquad\qquad \text{if} \;\; {\scriptstyle \lVert \sw^{(m)} \rVert^2+\sL(\cW, \sw^{(m)}) \le \gamma}
\end{cases}
\end{split}
\end{equation}
The gradient w.r.t $\cW$ just follows the conventional CNN based model plus the gradient that directly comes from the hidden layer supervision.

Next, we provide more discussions to and try to understand intuitively about our formulation, eqn. (\ref{eq:total2}).
For ease of reference, we write this objective function as
\begin{equation}
  F(\cW) \equiv \aP(\cW) + \aQ(\cW),
\end{equation}
where $\aP(\cW)\equiv \lVert \ow \rVert^2 +\cL(\cW, \ow)$ and $\aQ(\cW) \equiv  \sum_{m=1}^{M-1}  \alpha_m [\lVert \sw^{(m)} \rVert^2+\sL(\cW, \sw^{(m)})-\gamma]_+$.

\subsection{Stochastic Gradient Descent View}

We focus on the convergence advantage of DSN, instead of the regularization to the generalization aspect. In addition to the present problem in CNN where learned features are not always intuitive and discriminative \cite{Zeiler13}, the difficulty of training deep neural networks has been discussed \cite{Glorot10,Pascanu14}. As we can observe from eqn. (\ref{eq:Z}) and (\ref{eq:Q}), the change of the bottom layer weights get propagated through layers of functions, leading to exploding or vanishing gradients \cite{Pascanu14}. Various techniques and parameter tuning tricks have been proposed to better train deep neural networks, such as pre-training and dropout \cite{dropout}.
Here we provide a somewhat loose analysis to our proposed formulation, in a hope to understand its advantage in effectiveness. 

The objective function in deep neural networks is highly non-convex.
Here we make the following assumptions/observations: (1) the objective/energy function of DL observes a large ``flat'' area around the ``optimal'' solution where any result has a similar performance; locally we still assume a convex (or even $\lambda$-strongly convex) function whose optimization is often performed with stochastic gradient descent algorithm \cite{Bottou98}.    

The definition of $\lambda$-strongly convex is standard: A function $F(\cW)$ is $\lambda$-strongly convex if $\forall, \cW, \cW' \in \mathcal{W}$ and any subgradient $\cg$ at $\cW$,
\begin{equation}
  F(\cW') \ge F(\cW) + <\cg, \cW'-\cW> + \frac{\lambda}{2} \lVert \cW'-\cW \rVert^2,
\label{eq:overall}	
\end{equation}
and the update rule in Stochastic Gradient Descent (SGD) at step $t$ is $\cW_{t+1} = \Pi_{\mathcal{W}}(\cW_{t} - \eta_t \sg)$,
where $\eta_t=\Theta(1/t)$ refers to the step rate and $\Pi_{\mathcal{W}}$ helps to project onto the space of $\mathcal{W}$. Let $\cW^{*}$ be the optimum solution, upper bounds for $\cE[\lVert \cW_T-\cW^{*} \rVert^2]$ and 
$\cE[(F(\cW_T)-F(\cW^{*})^2]$ in \cite{Rakhlin12} for the strongly convex function, and $\cE[(F(\cW_T)-F(\cW^{*})^2]$ for convex function in \cite{Shamir13}.
Here we make an attempt to understand the convergence of eqn. (\ref{eq:overall}) w.r.t. $\cE[\lVert \cW_T-\cW^{*} \rVert^2]$, due to the presence of large area of flat function shown in Figure (\ref{fig:architecture}.b).
In \cite{Loh13}, a convergence rate is given for the M-estimators with locally convex function with compositional loss and regularization terms. Both terms in  eqn. (\ref{eq:overall}) here refer to the same class label prediction error, a reason for calling the second term as {\em companion objective}. Our motivation is two-fold: {\bf (1)} encourage the features learned at each layer to be directly discriminative for class label prediction, while keeping the ultimate goal of minimizing class label prediction at the output layer; {\bf (2)} alleviate the exploding and vanishing gradients problem as each layer now has a direct supervision from the ground truth labels. One might raise a concern that learning highly discriminative intermediate stage filters may not necessarily lead to the best prediction at the output layer.
An illustration can been seen in Figure (\ref{fig:architecture}.b).
Next, we give a loose theoretical analysis to our framework, which is also validated by comprehensive experimental studies with overwhelming advantages over the existing methods.

\begin{definition}
We name $\mathcal{S}_{\gamma}(F)=\{\cW: F(\cW) \le \gamma \}$ as the $\gamma$-feasible set for a function $F(\cW) \equiv \aP(\cW) + \aQ(\cW)$.
\end{definition}

First we show that a feasible solution for $\aQ(\cW)$ leads to a feasible one to $\aP(\cW)$. That is:
{\lemma \label{lm:1} $\forall m,m'=1..M-1, and \; m' > m$ \;
if $\; \lVert \sw^{(m)} \rVert^2+ \sL((\hat{\cW}^{(1)},..,\hat{\cW}^{(m)}), \sw^{(m)}) \le \gamma$ then there exists $(\hat{\cW}^{(1)},..,\hat{\cW}^{(m)},..,\hat{\cW}^{(m')})$ such that 
$\; \lVert \sw^{(m')} \rVert^2 + \sL((\hat{\cW}^{(1)},..,\hat{\cW}^{(m)}..,\hat{\cW}^{(m')}), \sw^{(m')}) \le \gamma$. \footnote{Note that we drop the $\cW^{(j)}, j>m$ since the filters above layer $m$ do not participate in the computation for the objective function of this layer.}
}
\begin{proof}

As we can see from an illustration of our network architecture shown in fig. (\ref{fig:architecture}.a), for $\forall \;(\hat{\cW}^{(1)},..,\hat{\cW}^{(m)})$ such that $\; \sL((\hat{\cW}^{(1)},..,\hat{\cW}^{(m)}), \sw^{(m)}) \le \gamma$. Then there is a trivial solution for the network for every layer $j>m$ up to $m'$, we let $\hat{\cW}^{(j)} = \mathbf{I}$ and $\sw^{(j)}=\sw^{(m)}$, meaning that the filters will be identity matrices. This results in $\; \sL((\hat{\cW}^{(1)},..,\hat{\cW}^{(m)}..,\hat{\cW}^{(m')}), \sw^{(m')}) \le \gamma$.  \hfill $\square$
\end{proof}

\begin{remark} Lemma \ref{lm:1} shows that a good solution for $\aQ(\cW)$ is also a good one for $\aP(\cW)$, but it may not be the case the other way around. That is: a $\cW$ that makes $\aP(\cW)$ small may not necessarily produce discriminative features for the hidden layers to have a small
$\aQ(\cW)$. However, $\aQ(\cW)$ can be viewed as a regularization term. Since $\aP(\cW)$ observes a very flat area near even zero on the training data and it is ultimately the test error that we really care about, we thus only focus on the $\cW$, $\cW^{\star}$, which makes both $\aQ(\cW)$ and $\aP(\cW)$ small. Therefore, it is not unreasonable to assume that $F(\cW)\equiv \aP(\cW) + \aQ(\cW)$ and $\aP(\cW)$ share the same optimal $\cW^{\star}$. 

 Let $\aP(\cW))$ and $\aP(\cW))$ be strongly convex around $\cW^{\star}$, $\lVert\cW'- \cW^{\star}\rVert^2 \le D$ and $\lVert\cW- \cW^{\star}\rVert^2 \le D$, with $\aP(\cW') \ge \aP(\cW) + <\gp, \cW'-\cW> + \frac{\lambda_1}{2} \lVert \cW'-\cW \rVert^2$ and $\aQ(\cW') \ge \aQ(\cW) + <\gq, \cW'-\cW> + \frac{\lambda_1}{2} \lVert \cW'-\cW \rVert^2$, where $\gp$  and $\gq$ are the subgradients for $\aP$ and $\aQ$ at $\cW$ respectively. It can be directly seen that $F(\cW)$ is also strongly convex and for subgradient $\gf$ of $F(\cW)$ at $\cW$, $\gf = \gp + \gq$.  
\end{remark}

{\lemma \label{lm:2}
Suppose $\cE[\lVert \hat{\gp}_t \rVert^2] \le G^2$ and $\cE[\lVert \hat{\gq}_t \rVert^2] \le G^2$, and we use the update rule of $\cW_{t+1} = \Pi_{\mathcal{W}}(\cW_{t} - \eta_t (\hat{\gp}_t+\hat{\gq}_t))$ where $\cE[\hat{\gp}_t]=\gp_t$ and $\cE[\hat{\gq}_t]=\gq_t$. If we use $\eta_t=1/(\lambda_1+\lambda_2)t$, then at time stamp $T$
\begin{equation}
\cE[\lVert \cW_T - \cW^{\star} \rVert^2] \le \frac{12G^2}{(\lambda_1+\lambda_2)^2 T}
\end{equation}
}
\begin{proof}
Since $F(\cW) = \aP(\cW) + \aQ(\cW)$, it can be directly seen that
\[
     F(\cW') \ge F(\cW) + <\gp+\gq, \cW'-\cW> + \frac{\lambda_1+\lambda_2}{2} \lVert \cW'-\cW \rVert^2.
\]
Based on lemma 1 in \cite{Rakhlin12}, this upper bound directly holds. \hfill $\square$

\end{proof}

{\lemma \label{lm:3}
Following the assumptions in lemma \ref{lm:2}, but now we assume $\eta_t=1/t$ since $\lambda_1$ and $\lambda_2$ are not always readily available, then started from $\lVert \cW_1-\cW^{\star} \rVert^2 \le D$ the convergence rate is bounded by
\begin{equation}
\cE[\lVert \cW_T - \cWs \rVert^2] \le e^{-2\lambda (\ln T + 0.578)} D + (T-1) e^{-2\lambda \ln(T-1)} G^2
\end{equation}
}
\begin{proof} 
Let $\lambda = \lambda_1 + \lambda_2$, we have
\[
   F(\cWs )- F(\cW_t) \ge \;\;<\gf_t, \cWs - \cW_t> + \frac{\lambda}{2} \lVert \cWs - \cW_t \rVert^2, \;\;and
\]

\[
   F(\cWs) - F(\cW_t) \ge \frac{\lambda}{2} \lVert \cW_t- \cWs \rVert^2.
\]
Thus,
\[
   <\gf_t, \cW_t - \cWs> \;\; \ge \lambda \lVert \cW_t- \cWs \rVert^2
\]
Therefore, with $\eta_t=1/t$,
\begin{eqnarray}
 \cE[\lVert \cW_{t+1} - \cWs \rVert^2] &=& \cE[\lVert \Pi_{\mathcal{W}}(\cW_{t} - \eta_t \hat{\gf}_t) - \cWs \rVert^2] \nonumber \\
 &\le& \cE[\lVert \cW_{t} - \eta_t \hat{\gf}_t - \cWs \rVert^2] \nonumber \\
&=& \cE[\lVert \cW_{t} - \cWs \rVert^2] - 2 \eta_t \cE[<\gf_t, \cW_t-\cWs>] + \eta_t \cE[\lVert \hat{\gf}_t \rVert^2] \nonumber \\
&\le& (1-2\lambda/t) \cE[\lVert \cW_{t} - \cWs \rVert^2] + G^2/t^2
\end{eqnarray}
With $2\lambda/t$ being small, we have $1-2\lambda/t \approx e^{-2\lambda/t}.$
\begin{eqnarray}
\cE[\lVert \cW_T - \cWs \rVert^2] &\le& e^{-2\lambda(\frac{1}{1}+\frac{1}{2}+,..,\frac{1}{T})} D + \sum_{t=1}^{T-1} \frac{G^2}{t^2} e^{-2\lambda(\frac{1}{t}+\frac{1}{t+1}+,..,\frac{1}{T-1})} \nonumber \\
&=& e^{-2\lambda (\ln T + 0.578)} D + G^2 \sum_{t=1}^{T-1} e^{-2 \ln(t)- 2 \lambda(\ln(T-1)-2\lambda \ln(t)} \nonumber \\
&\le& e^{-2\lambda (\ln T + 0.578)} D + (T-1) e^{-2\lambda \ln(T-1)} G^2 \nonumber \quad\quad\quad\quad\quad\quad\quad\quad\quad  \square
\end{eqnarray}
\end{proof}

{\theor Let $\aP(\cW)$ be $\lambda_1$-strongly convex and $\aQ(\cW)$ be $\lambda_2$-strongly convex near optimal $\cWs$ and denote $\cW_T^{(F)}$ and $\cW_T^{(\aP)}$ as the solution after $T$ iterations when applying SGD on $F(\cW)$ and $\aP(\cW)$ respectively. Then our deeply supervised framework in eqn. (\ref{eq:total2}) improves the the speed over using top layer only by
$\frac{\cE[\lVert \cW_T^{(\aP)}- \cWs \rVert^2]}{\cE[\lVert \cW_T^{(F)} - \cWs \rVert^2]} = \Theta (1+\frac{\lambda_2^2}{\lambda_1^2}), \; when \; \eta_t=1/\lambda t, \quad and,$
$\frac{\cE[\lVert \cW_T^{(\aP)}- \cWs \rVert^2]}{\cE[\lVert \cW_T^{(F)} - \cWs \rVert^2]} = \Theta (e^{\ln(T) \lambda_2}), \; when \; \eta_t=1/t.$
}
\begin{proof}

Lemma \ref{lm:1} shows the compatibility of the companion objective of $\aQ$ w.r.t the output objective $\aP$. The first equation can be directly derived from lemma \ref{lm:2} and the second equation can be seen from lemma \ref{lm:3}. In general $\lambda_2 \gg \lambda_1$ which leads to a great improvement in convergence speed and the constraints in each hidden layer also helps to learning filters which are directly discriminative.
\hfill $\square$
\end{proof}

\vspace{-1mm}
\section{Experiments}
\vspace{-2mm}
We evaluate the proposed DSN method on four standard benchmark datasets: MNIST, CIFAR-10, CIFAR-100 and SVHN. 
We follow a common training protocol used by Krizhevsky et al. \cite{imagenet} in all experiments. 
We use SGD solver with mini-batch size of $128$ at a fixed constant momentum value of $0.9$. Initial value for learning rate and weight decay factor is determined based on the validation set. For a fair comparison and clear illustration of the effectiveness of DSN, we match the complexity of our model with that in network architectures used in \cite{NIN} and \cite{maxout} to have a comparable number of parameters. We also incorporate two dropout layers with dropout rate at $0.5$. Companion objective at the convolutional layers is imposed to backpropagate the classification error guidance to the underlying convolutional layers. Learning rates are annealed during training by a factor of $20$ according to an epoch schedule determined on the validation set. The proposed DSN framework is not difficult to train and there are no particular engineering tricks adopted. Our system is built on top of widely used Caffe infrastructure \cite{caffe}.
For the network architecture setup, we adopted the mlpconv layer and global averaged pooling scheme introduced in \cite{NIN}. 
DSN can be equipped with different types of loss functions, such as Softmax and SVM. We show performance boost of DSN-SVM and DSN-Softmax over CNN-SVM and CNN-Softmax respectively (see Figure (\ref{fig:mnist}.a)). The performance gain is more evident in presence of small training data (see Figure (\ref{fig:mnist}.b)); this might partially ease the burden of requiring large training data for DL. Overall, we observe state-of-the-art classification error in all four datasets (without data augmentation), $0.39\%$ for MINIST, $9.78\%$ for CIFAR-10, $34.57\%$ for CIFAR-100, and $1.92\%$ for SVHN ($8.22\%$ for CIFAR-10 with data augmentation). All results are achieved without using averaging \cite{mulCOL}, which is not exclusive to our method. Figure (\ref{fig:visualization}) gives an illustration of some learned features.

\subsection{MNIST}
\vspace{-1mm}
\begin{figure}[ht]
\begin{center}
\begin{tabular}{ccc}
\hspace{-5mm}			\includegraphics[width=0.3\linewidth, height=0.3\linewidth]{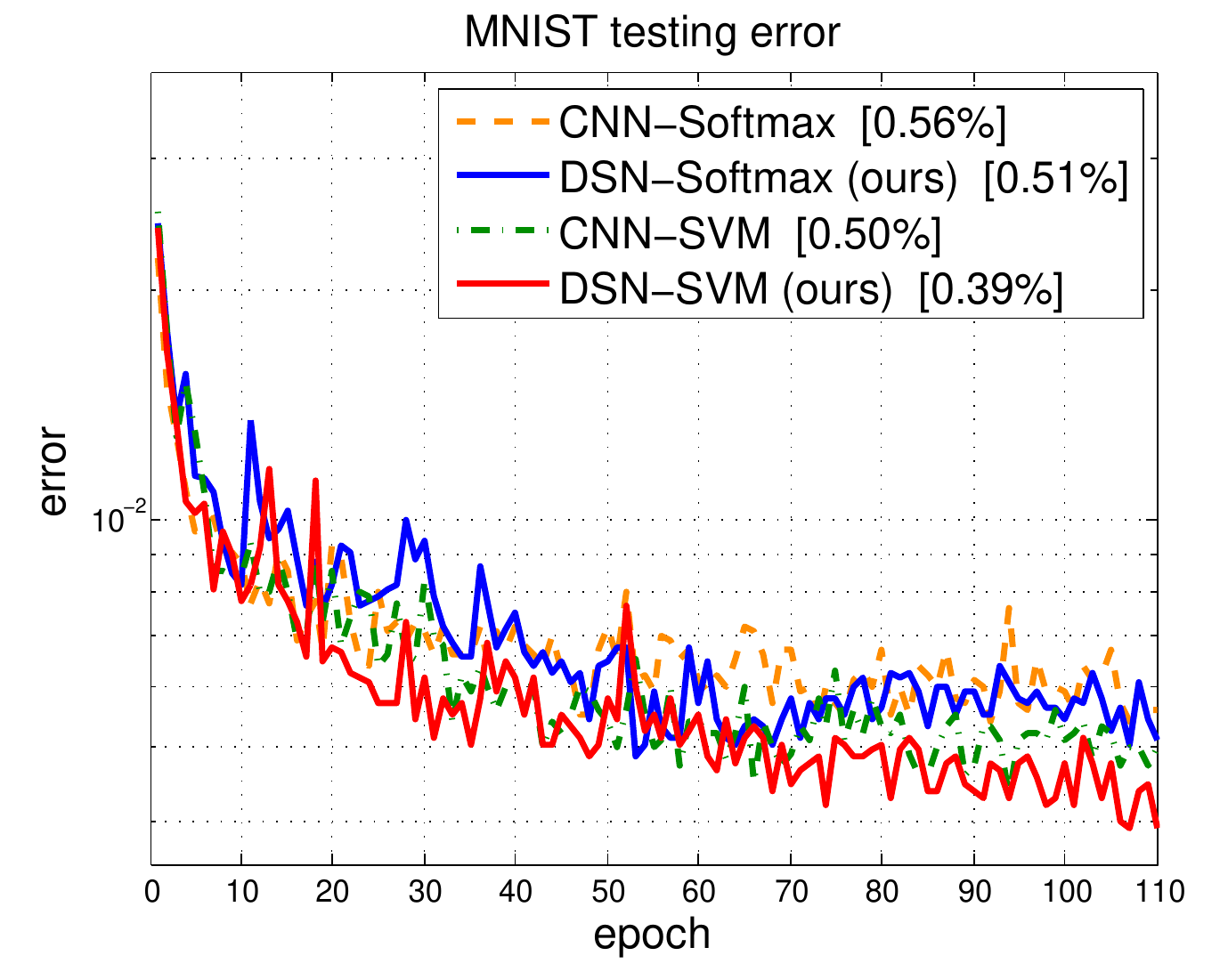} &
\hspace{-5mm}			\includegraphics[width=0.3\linewidth]{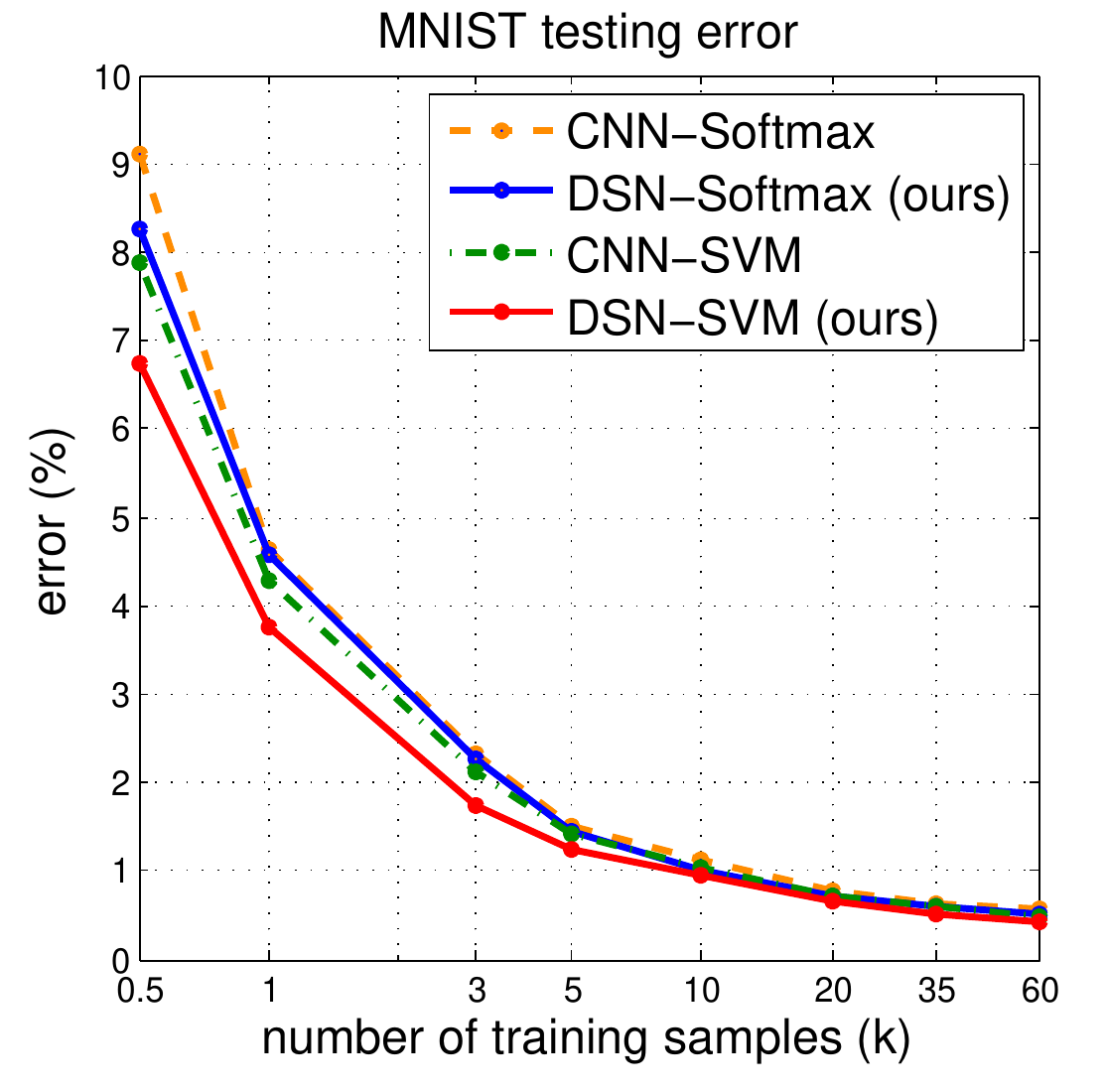} &
\hspace{-5mm}			\includegraphics[width=0.3\linewidth]{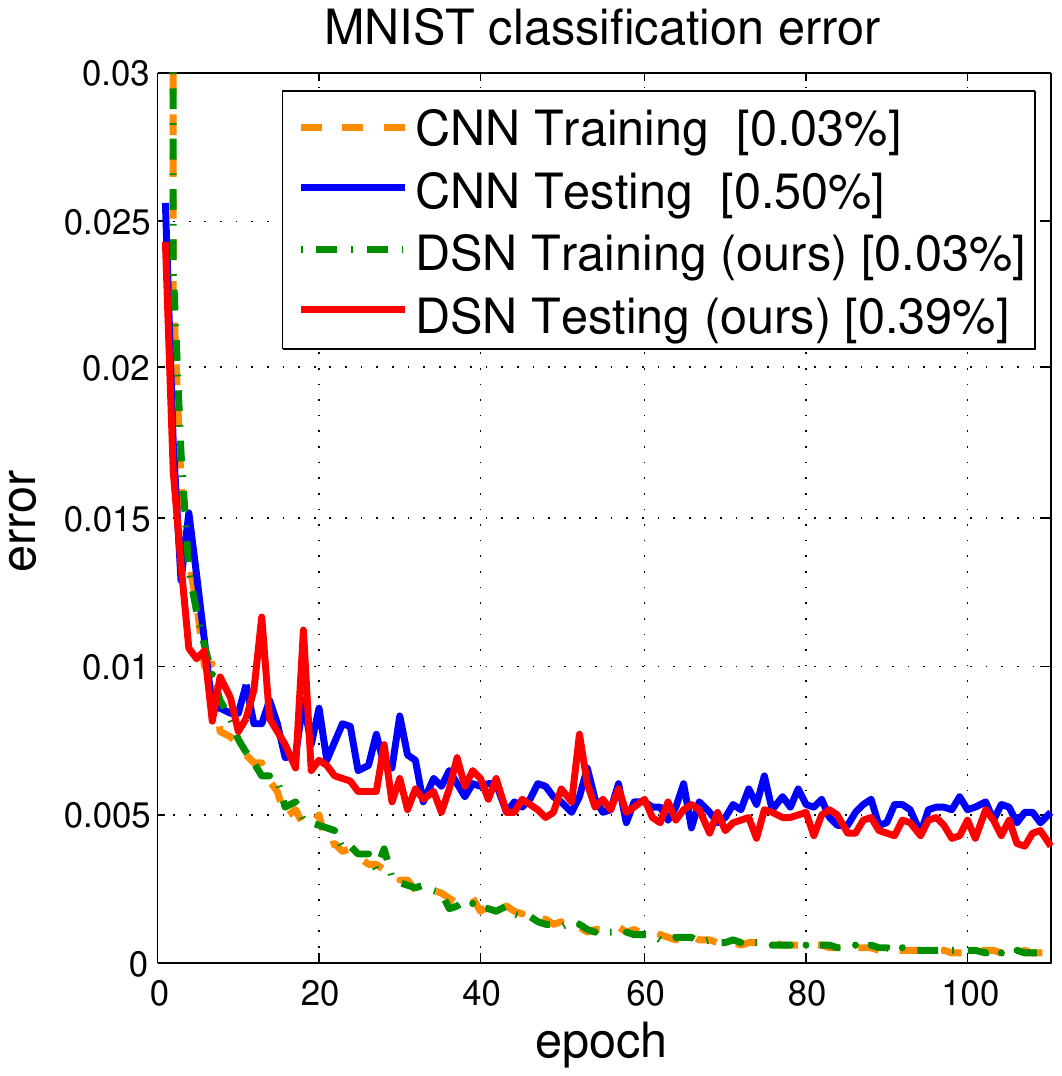}\\
			(a) & (b) & (c) \\
\end{tabular}
\vspace{-3mm}
\caption{Classification error on MNIST test. (a) shows test error of competing methods; (b) shows test error w.r.t. the training sample size. (c) training and testing error comparison.}
\label{fig:mnist}
\end{center}
\vspace{-5mm}
\end{figure}

We first validate the effectiveness of the proposed DSN on the MNIST handwritten digits classification task \cite{Lecun98}, a widely and extensively adopted benchmark in machine learning. MNIST dataset consists of images of 10 different classes (0 to 9) of size $28 \times 28$ with 60,000 training samples and 10,000 test samples.  Figure \ref{fig:mnist}(a) and (b) show results from four methods, namely: (1) conventional CNN with softmax loss (CNN-Softmax), (2) the proposed DSN with softmax loss (DSN-Softmax), (3) CNN with max-margin objective (CNN-SVM) , and (4) the proposed DSN with max-margin objective (DSN-SVM). DSN-Softmax and DSN-SVM outperform both their competing CNN algorithms (DSN-SVM shows classification error of $0.39\%$ under a single model without data whitening and augmentation). Figure \ref{fig:mnist}(b) shows classification error of the competing methods when trained w.r.t. varying sizes of training samples ($26\%$ gain of DSN-SVM over CNN-Softmax at $500$ samples. Figure \ref{fig:mnist}(c) shows a comparison of generalization error between CNN and DSN. 

\begin{table}[!htp]
\centering
\caption{MNIST classification result (without using data augmentation).}
	\small
	\begin{tabular}[t]{lc}
	\multicolumn{1}{c}{\bf Method}  &\multicolumn{1}{c}{\bf Error(\%)} 
	\\ \hline
	CNN \cite{objREC} & 0.53\\
	Stochastic Pooling \cite{SPool} & 0.47 \\
	Network in Network \cite{NIN} & 0.47 \\
	Maxout Networks\cite{maxout} & 0.45 \\
	\textbf{DSN (ours)} & \textbf{0.39} \\
	\end{tabular}
	\captionsetup{labelformat=empty}
\end{table}

\vspace{-3mm}
\subsection{CIFAR-10 and CIFAR-100}

\begin{figure}
\vspace{-2mm}
\begin{center}
\begin{tabular}{c}
\includegraphics[width=90mm]{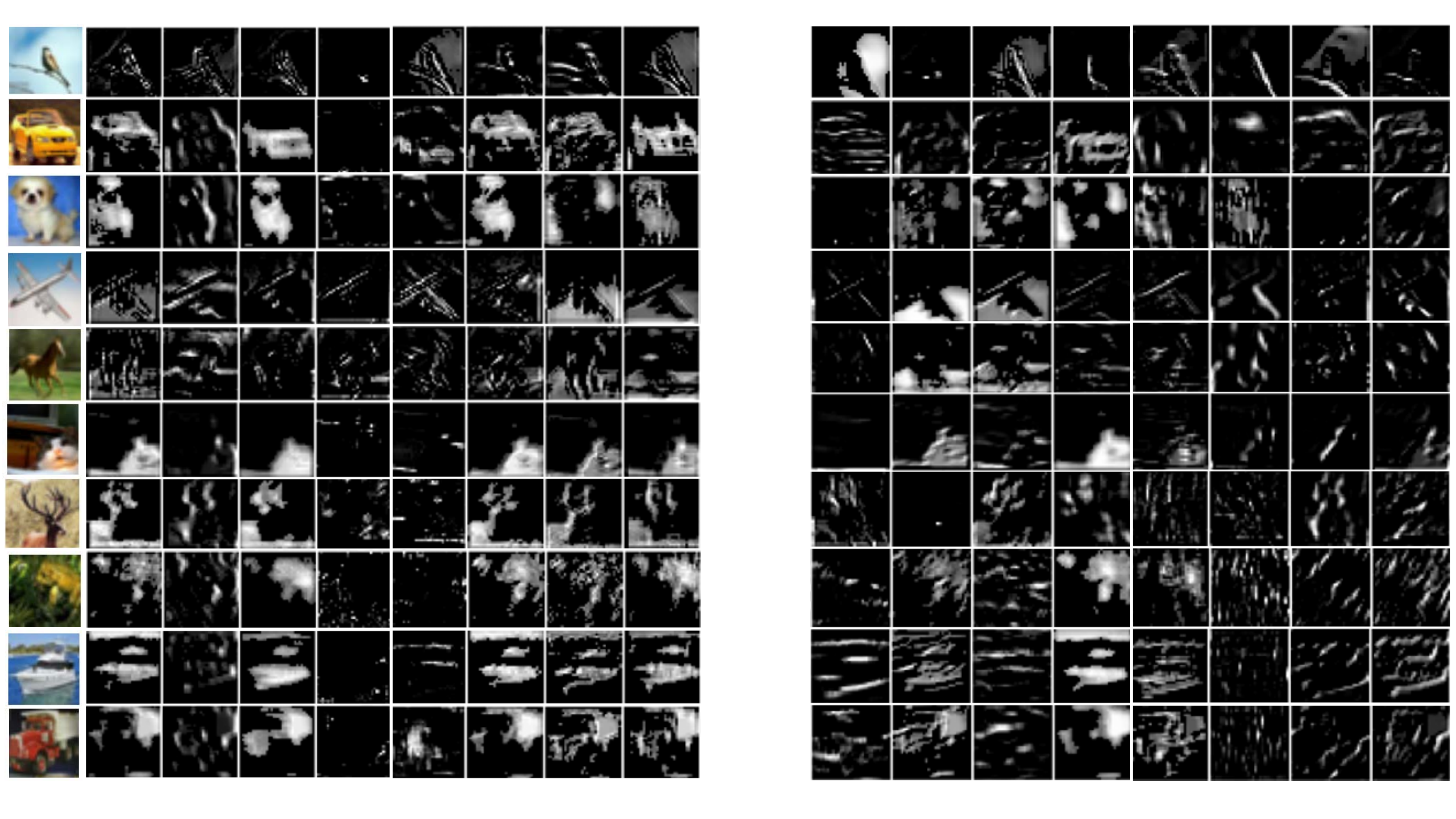} \\
(a) by DSN \hspace{15mm}  (b) by CNN \\
\end{tabular}
\vspace{-3mm}
\caption{Visualization of the feature maps learned in the convolutional layer.}
\label{fig:visualization}
\end{center}
\vspace{-5mm}
\end{figure}

CIFAR-10 dataset consists of $32 \times 32$ color images. A total number of 60,000 images are split into 50,000 training and 10,000 testing images. The dataset is preprocessed by global contrast normalization. To compare our results with the previous state-of-the-art, in this case, we also augmented the dataset by zero padding 4 pixels on each side, then do corner cropping and random flipping on the fly during training.
No model averaging is done at the test phase and we only crop the center of a test sample. 
Table (\ref{cifar100-table}) shows our result. Our DSN model achieved an error rates of $9.78\%$ without data augmentation and $8.22\%$ with data agumentation (the best known result to our knowledge).

DSN also provides added robustness to hyperparameter choice, in that the early layers are guided with direct classification loss, leading to a faster convergence rate and relieved burden on heavy hyperparameter tuning. 
We also compared the gradients in DSN and those in CNN, observing $4.55$ times greater gradient variance of DSN over CNN in the  first convolutional layer. This is consistent with an observation in \cite{maxout}, and the assumptions and motivations we make in this work. 
To see what the features have been learned in DSN vs. CNN, we select one example image from each of the ten categories of CIFAR-10 dataset, run one forward pass, and show the feature maps learned from the first (bottom) convolutional layer in Figure (\ref{fig:visualization}). Only the top 30\% activations are shown in each of the feature maps. Feature maps learned by DSN show to be more intuitive than those by CNN.

\begin{table*}[!htp]
\begin{tabular}{cc}

CIFAR-10 classification error & CIFAR-100 classification error\\

\begin{minipage}[b]{0.49\linewidth}
\begin{center}
\begin{tabular}{ll}
\multicolumn{1}{c}{\bf Method}  &\multicolumn{1}{c}{\bf Error(\%)} 
\\ \hline
\tiny{No Data Augmentation}
\\ \hline
Stochastic Pooling \cite{SPool} &15.13 \\
Maxout Networks   \cite{maxout}     &11.68 \\
Network in Network \cite{NIN} &10.41 \\
\textbf{DSN (ours)}   & \textbf{9.78} \\
\hline
\tiny{With Data Augmentation}
\\ \hline
Maxout Networks \cite{maxout} &9.38\\
DropConnect \cite{dropcon}   &9.32 \\
Network in Network \cite{NIN}&8.81 \\
\textbf{DSN (ours)}   			 &\textbf{8.22} \\
\end{tabular}
\end{center}
\end{minipage}
&
\begin{minipage}[b]{0.49\linewidth}
\begin{center}
\begin{tabular}{ll}
\multicolumn{1}{c}{\bf Method}  &\multicolumn{1}{c}{\bf Error(\%)} 
\\ \hline
Stochastic Pooling   \cite{SPool}  &42.51 \\
Maxout Networks  \cite{maxout}      &38.57 \\
Tree based Priors  \cite{tree}   & 36.85 \\
Network in Network  \cite{NIN} &35.68 \\
\textbf{DSN (ours)}   			 &\textbf{34.57} \\
\end{tabular}
\end{center}
\end{minipage} \\
\end{tabular}
\caption{Method comparison on CIFAR-10 and CIFAR-100 test data.}
\label{cifar100-table}
\end{table*}

CIFAR-100 dataset is similar to CIFAR-10 dataset, except that it has 100 classes. The number of images for each class is then $500$ instead of $5,000$ as in CIFAR-10,  which makes the classification task more challenging. We use the same network settings as in CIFAR-10. Table (\ref{cifar100-table}) shows previous best results and $34.57\%$ is reported by DSN. The performance boost consistently shown on both CIFAR-10 and CIFAR-100 again demonstrates the advantage of the DSN method.

\vspace{-3mm}
\subsection{Street View House Numbers}
\vspace{-3mm}

\begin{table}[!htp]
\vspace{-5mm}
			\centering
			\begin{tabular}{ll}
			\multicolumn{1}{c}{\bf Method}  &\multicolumn{1}{c}{\bf Error(\%)} 
			\\ \hline
			Stochastic Pooling  \cite{SPool} &2.80 \\
			Maxout Networks     \cite{maxout} &2.47 \\
			Network in Network  \cite{NIN} &2.35 \\
			Dropconnect    \cite{dropcon}    & 1.94 \\
			\textbf{DSN (ours)}   			 &\textbf{1.92} \\
			\end{tabular}
  \caption{SVHN classification error.}
	\label{svhn-table}
\end{table}

Street View House Numbers (SVHN) dataset consists of $73,257$ digits for training, $26,032$ digits for testing, and $53,1131$ extra training samples on $32 \times 32$ color images. We followed the previous works for data preparation, namely: we select 400 samples per class from the training set and 200 samples per class from the extra set. The remaining 598,388 images are used for training. We followed \cite{maxout} to preprocess the dataset by Local Contrast Normalization (LCN). We do not do data augmentation in training and use only a single model in testing. Table \ref{svhn-table} shows recent comparable results.  Note that Dropconnect \cite{dropcon} uses data augmentation and multiple model voting.

\vspace{-2mm}
\section{Conclusions}
\vspace{-1mm}
In this paper, we have presented a new formulation, deeply-supervised nets (DSN), attempting to make a more transparent learning process for deep learning. Evident performance enhancement over existing approaches has been obtained. A stochastic gradient view also sheds light to the understanding of our formulation.

\section{Acknowledgments}
This work is supported by NSF award IIS-1216528 (IIS-1360566) and NSF award IIS-0844566 (IIS-1360568). We thank Min Lin, Naiyan Wang, Baoyuan Wang, Jingdong Wang, Liwei Wang, and David Wipf for help discussions. We are greatful for the generous donation of the GPUs by NVIDIA.

\small{
\bibliographystyle{ieee}
\bibliography{egbib}
}

\end{document}